\documentclass[letterpaper,onecolumn,10pt]{article}
\usepackage[utf8]{inputenc}
\usepackage{url}
\usepackage{graphicx}
\usepackage{makecell}
\usepackage{pdfpages}
\usepackage{booktabs}
\usepackage{hyperref}

\usepackage{array}
\newcolumntype{L}[1]{>{\raggedright\let\newline\\\arraybackslash\hspace{0pt}}m{#1}}
\newcolumntype{C}[1]{>{\centering\let\newline\\\arraybackslash\hspace{0pt}}m{#1}}
\newcolumntype{R}[1]{>{\raggedleft\let\newline\\\arraybackslash\hspace{0pt}}m{#1}}

\usepackage{geometry}
\usepackage{caption}  % Use for \captionof(*) command

\usepackage{pdflscape}

\usepackage{etoolbox}
\makeatletter
\newcommand*{\AM@pagecommandstar}{}
\define@key{pdfpages}{pagecommand*}{\def\AM@pagecommandstar{#1}}
\patchcmd{\AM@output}{\begingroup\AM@pagecommand\endgroup}
{\ifthenelse{\boolean{AM@firstpage}}{\begingroup\AM@pagecommandstar\endgroup}{\begingroup\AM@pagecommand\endgroup}}{}{} % Patch to use new option
\patchcmd{\AM@split@optionsii}{\equal{pagecommand}{\AM@temp}\or}
{\equal{pagecommand}{\AM@temp}\or\equal{pagecommand*}{\AM@temp}\or}{}{}
\makeatother

\newcommand*{\affaddr}[1]{#1} % No op here. Customize it for different styles.
\newcommand*{\affmark}[1][*]{\textsuperscript{#1}}
\newcommand*{\email}[1]{\texttt{#1}}

\begin{document}
% Don't want date printed
\date{}
% Make title large and bold
\title{\Large\bfseries Replication study: Development and validation of a deep learning algorithm for detection of diabetic retinopathy in retinal fundus photographs}

% Target typesetting:
%
% Author A, Author B, Author C, Author D and Author E
%        A,B,C Department of Computer Science
%       D,E Department of Mechanical Engineering
%          Email A,B,C,D,E @university.edu
%                  Latex University

\author{%
Mike Voets\affmark[1], Kajsa Møllersen\affmark[2], Lars Ailo Bongo\affmark[1]\\
\affaddr{\affmark[1]Department of Computer Science}\\
\affaddr{\affmark[2]Department of Community Medicine}\\
\affaddr{UiT - The Arctic University of Norway}\\
\email{mwhg.voets@gmail.com,\{kajsa.mollersen,lars.ailo.bongo\}@uit.no}
}

\maketitle

\begin{abstract}
    Replication studies are essential for validation of new methods, and are crucial to maintain the high standards of scientific publications, and to use the results in practice. 

    We have attempted to replicate the main method in \textit{Development and validation of a deep learning algorithm for detection of diabetic retinopathy in retinal fundus photographs} published in JAMA 2016; 316(22)\cite{Gulshan2016}. We re-implemented the method since the source code is not available, and we used publicly available data sets.
    
    The original study used non-public fundus images from EyePACS and three hospitals in India for training. We used a different EyePACS data set from Kaggle. The original study used the benchmark data set Messidor-2 to evaluate the algorithm's performance. We used the same data set. In the original study, ophthalmologists re-graded all images for diabetic retinopathy, macular edema, and image gradability. There was one diabetic retinopathy grade per image for our data sets, and we assessed image gradability ourselves. The original study did not provide hyper-parameter settings. But some of these were later published.

    We were not able to replicate the original study, due to insufficient level of detail in the method description. Our best effort of replication resulted in an algorithm that was not able to reproduce the results of the original study. Our algorithm's area under the receiver operating characteristic curve (AUC) of 0.94 on the Kaggle EyePACS test set and 0.80 on Messidor-2 did not come close to the reported AUC of 0.99 on both test sets in the original study. This may be caused by the use of a single grade per image, different data, or different not described hyper-parameters. We conducted experiments with various normalization methods, and found that training by normalizing the images to a [--1, 1] range gave the best results for this replication.

    This study shows the challenges of replicating deep learning method, and the need for more replication studies to validate deep learning methods, especially for medical image analysis. Our source code and instructions are available at: \\
    \url{https://github.com/mikevoets/jama16-retina-replication}
\end{abstract}

\newpage

\section{Introduction}

Being able to replicate a scientific paper by strictly following the described methods is a cornerstone of science. Replicability is essential for the development of medical technologies based on published results. However, there is an emerging concern that many studies are not replicable, raised for bio-medical research \cite{eLife:20176}, computational sciences \cite{Nature:496, Nature:533}, and recently for machine learning \cite{Hutson2018}.

The terms replicate and reproduce are often used without precise definition, and sometimes referring to the same concept. We will distinguish the two in the scientific context by defining replication as repeating the method as described and reproducing as obtaining the result as reported. The scientific standard for publication is to describe the method with sufficient detail for the study to be repeated, and a replication study is then the attempt to repeat the study as described, not as it might have been conducted. To balance the level of detail and readability of a manuscript, a replication attempt must follow the standard procedures of the field when details about the method are missing.

If the data that produced the reported results are available, the replicated method should reproduce the original results, and any deviation points towards a lack of detail in the description of the method, assuming that the replication is conducted correctly. 
If the data are not available, deviations in the results can be due to either insufficient description of the method, or differences in the data, and it will not be possible to separate the two sources of deviation. 
This should not prevent replication studies from being conducted, even if the conclusions regarding replicability are less definite than if the data were available. 

Deep learning has become a hot topic within machine learning due to its promising performance of finding patterns in large data sets. There are dozens of libraries that make deep learning methods easily available for any developer. This has consequently led to an increase of published articles that demonstrate the feasibility of applying deep learning in practice, particularly for image classification \cite{Litjens2017, Ching2018}. However, there is an emerging need to show that studies are replicable, and hence be used to develop new medical analysis solutions. Ideally, the data set and the source code are published, so that other researchers can verify the results by using the same or other data. However, this is not always practical, for example for sensitive data, or for methods with commercial value \cite{Nature:533, Hutson2018}.

In this study, we make an assessment on the replicability of a deep learning method. We have chosen to attempt to replicate the main method from \textit{Development and validation of a deep learning algorithm for detection of diabetic retinopathy in retinal fundus photographs}, published in JAMA 2016; 316(22)\cite{Gulshan2016}. As of April 2018, this article had been cited 350 times \cite{GoogleScholar:Citations}. We chose to replicate this study because it is a well-known and high-impact study within the medical field, the source code has not been published, and there are as far as we know not any others who have attempted to replicate this study.

Deep learning methods have been used in other publications for detection of diabetic retinopathy \cite{Ting2017DevelopmentDiabetes, Abramoff2016ImprovedLearning, Gargeya2017AutomatedLearning} with high performance. Direct comparison between these algorithms is not possible, since they use different data sets for evaluation of performance. However, these papers confirm the main findings of \cite{Gulshan2016}; that deep learning can be used to automatically detect diabetic retinopathy. We assess \cite{Gulshan2016} to be most promising regarding replicability, due to its enhanced focus on method and more detailed descriptions. 

The original study describes an algorithm (hereby referred to as the original algorithm) for detection of referable diabetic retinopathy (rDR) in retinal fundus photographs. The algorithm is trained and validated using 118 419 fundus images retrieved from EyePACS and from three eye hospitals in India. The original algorithm's performance was evaluated on 2 test sets, and achieved an area under the receiver operating characteristic curve (AUC) for detecting rDR of 0.99 for both the EyePACS-1 and the Messidor-2 test sets. Two operating points were selected for high sensitivity and specificity. The operating point for high specificity had 90.3\% and 87.0\% sensitivity and 98.1\% and 98.5\% specificity for the EyePACS-1 and Messidor-2 test sets, whereas the operating point for high sensitivity had 97.5\% and 96.1\% sensitivity and 93.4\% and 93.9\% specificity, respectively.

To assess replicability of the method used to develop the original algorithm for detection of rDR, we used similar images from a publicly available EyePACS data set for training and validation, and we used a subset from the EyePACS data set and images from the public Messidor-2 data set for performance evaluation. We had to find validation hyper-parameters and the image normalization method ourselves, because they were not described in the original study. Our objective is to compare the performance of the original rDR detection algorithm to our result algorithm after trying to replicate, taking into account potential deviations in the data sets, having fewer grades, and potential differences in normalization methods and other hyper-parameter settings.

We were not able to replicate the original study, due to insufficient level of detail in the method description. Our best effort of replication resulted in an algorithm that was not able to reproduce the results of the original study. Our algorithm's AUC for detecting rDR for our EyePACS and Messidor-2 test sets were 0.94 and 0.80, respectively. The operating point for high specificity had 83.4\% and 67.9\% sensitivity and 90.1\% and 76.4\% specificity for our EyePACS and Messidor-2 test sets, and the operating point for high sensitivity had 89.9\% and 73.7\% sensitivity and 83.8\% and 69.7\% specificity. The results can differ for four reasons. First, we used public retinal images with only one grade per image, whereas in the original study the non-public retinal images were re-graded multiple times. Second, the now published list of hyper-parameters used in the original study \cite{Krause2018} lack details regarding the normalization method and the validation procedure used, so the original algorithm may have been tuned better. Third, there might be errors in the original study or methodology. The last possible reason is that we may have done something wrong with replicating the method by having misinterpreted the methodology. We do not know for sure which of the four reasons has led to our considerably worse performance.

We do not believe our results invalidate the main findings of the original study. However, our result gives a general insight into the challenges of replicating studies that do not use publicly available data and publish source code, and it motivates the need for additional replication studies in deep learning. We have published our source code with instructions for how to use it with public data. This gives others the opportunity to improve upon the attempted replication.

\section{Methods}

\subsection{Data sets}
The data sets consist of images of the retinal fundus acquired for diabetic retinopathy screening. Any other information regarding the patient is not part of the data sets. Each image is graded according to severity of symptoms (see Section~\ref{sec:data-grading}).

The original study obtained 128 175 retinal fundus images from EyePACS in the US and from three eye hospitals in India. 118 419 macula-centered images from this data set were used for algorithm training and validation (referred to as \emph{development set}, divided into \emph{training} and \emph{tuning set} in the original study).
To evaluate the performance of the algorithm, the original study used two data sets (referred to as {\it validation sets} in the original study). 
For evaluating an algorithm's performance, the term test set is commonly used. The first test set was a randomly sampled set of 9963 images retrieved at EyePACS screening sites between May 2015 and October 2015. 
The second test set was the publicly available Messidor-2 data set \cite{Decenciere2014, Quellec2008}, consisting of 1748 images. We provide an overview of the differences in image distribution used in our replication study compared with the original study in Figure~\ref{fig:data-distribution}.

We obtained images for training, validation and testing from two sources: EyePACS from a Kaggle competition \cite{Kaggle:Main}, and the Messidor-2 set that was used in the original study. The Messidor-2 set is a benchmark for algorithms that detect diabetic retinopathy. We randomly sampled the Kaggle EyePACS data set consisting of 88 702 images into a training and validation set of 57 146 images and a test set of 8790 images. The leftover images were mostly images graded as having no diabetic retinopathy and were not used for training the algorithm. The reason for the number of images in our training and validation set is to keep the same balance for the binary rDR class as in the original study’s training and validation set. Our EyePACS test set has an identical amount of images and balance for the binary rDR class as in the original study’s EyePACS test set. We used all 1748 images from the Messidor-2 test set.

\subsection{Grading}
\label{sec:data-grading}

The images used for the algorithm training and testing in the original study were all graded by ophthalmologists for image quality (gradability), the presence of diabetic retinopathy, and macular edema. We did not have grades for macular edema for all our images, so we did not train our algorithm to detect macular edema.

\begin{figure}
  \centering
    \includegraphics[width=0.45\textwidth]{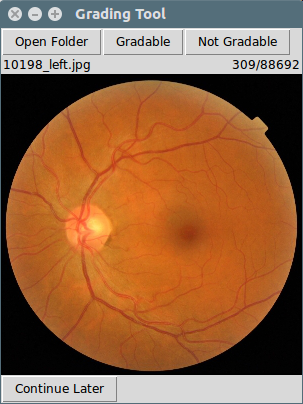}
  \caption{Screenshot of grading tool used to assess gradability for all images.\label{fig:grading-tool}}
\end{figure}

\newgeometry{left=0.9cm,right=0.9cm,bottom=1cm,footskip=1em,includefoot}

\includepdf[
    pages=-,
    scale=0.80,
    pagecommand*={\thispagestyle{plain}\null\vfill\captionof{figure}{Data set distribution in original study vs. this replication study. \label{fig:data-distribution}}}
]{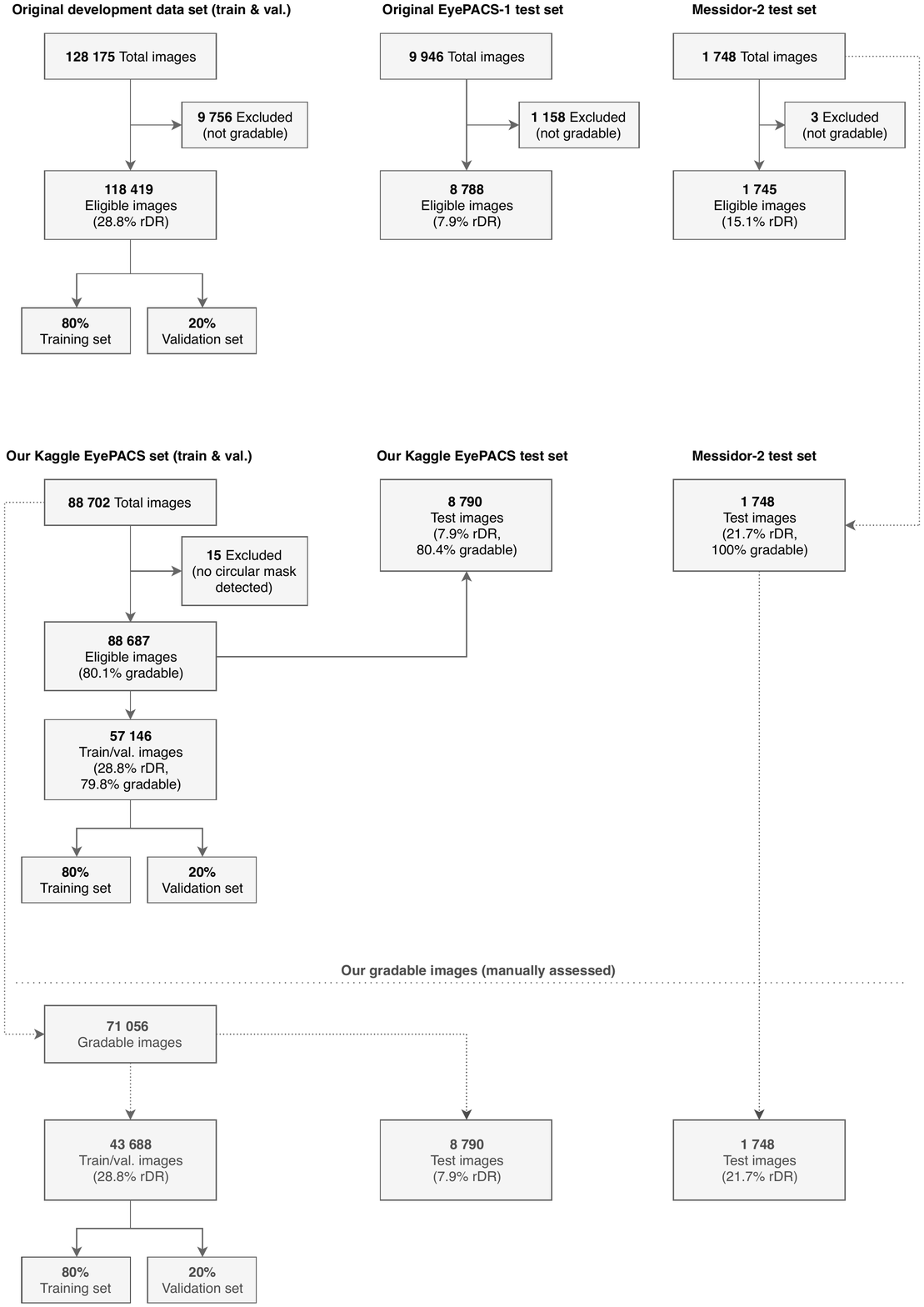}

\restoregeometry

Kaggle \cite{Kaggle:Data} describes that some of the images in their EyePACS distribution may consist of noise, contain artifacts, be out of focus, or be over- or underexposed. \cite{Rakhlin2017} states further that 75\% of the EyePACS images via Kaggle are estimated gradable. For this study one of the authors (MV) graded all Kaggle and Messidor-Original images on their image quality with a simple grading tool (Figure \ref{fig:grading-tool}). MV is not a licensed ophthalmologist, but we assume fundus image quality can be reliably graded by non-experts. We used the “Grading Instructions” in the Supplement of the original study to assess image quality. We publish the image quality grades with the source code. Images of at least adequate quality were considered gradable.

In the original study, diabetic retinopathy was graded according to the International Clinical Diabetic Retinopathy scale \cite{Icoph:2010}, with no, mild, moderate, severe or proliferative severity.

\begin{figure}
  \label{fig:ungradable-images}
  \centering
    \includegraphics[width=1.0\textwidth]{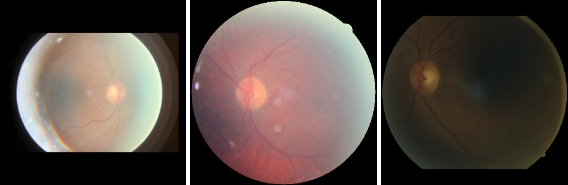}
  \caption{Examples of ungradable images because they are either out of focus, under-, or overexposed.}
\end{figure}

The Kaggle EyePACS set had been graded by one clinician for the presence of diabetic retinopathy using the same international scaling standard as used in the original study. We have thus only one diagnosis grade for each image. Kaggle does not give more information about where the data is from. The Messidor-2 test set and its diabetic retinopathy grades were made available by Ambramoff \cite{AbramoffDRGrades}.

\subsection{Algorithm training}

The objective of this study is to assess replicability of the original study. We try to replicate the method by following the original study's methodology as accurately as possible. As in the original study, our algorithm is created through deep learning, which involves a procedure of training a neural network to perform the task of classifying images. We trained the algorithm with the same neural network architecture as in the original study: the InceptionV3 model proposed by Szegedy et al \cite{Szegedy2015}. This neural network consists of a range of convolutional layers that transforms pixel intensities to local features before converting them into global features.

The fundus images from both training and test sets were preprocessed as described by the original study’s protocol for preprocessing. In all images the center and radius of the each fundus were located and resized such that each image gets a height and width of 299 pixels, with the fundus center in the middle of the image. A later article reports a list of data augmentation and training hyper-parameters for the trained algorithm in the original study \cite{Krause2018}. We applied the same data augmentation settings in our image preprocessing procedure.

The original study used distributed stochastic gradient descent proposed by Dean et al \cite{Dean2012} as the optimization function for training the parameters (i.e. weights) of the neural network. This implies that their neural network was trained in parallel, although the paper does not describe it. We did not conduct any distributed training for our replica neural network. According to the hyper-parameters published in \cite{Krause2018}, the optimization method that was used in the original study was RMSProp. Therefore, we used RMSProp as our optimization procedure. The hyper-parameter list specifies a learning rate of 0.001, so we used this same learning rate for our algorithm training. We furthermore applied the same weight decay of $4*10^{-5}$.

As in the original study, we used batch normalization layers \cite{Ioffe2015} after each convolutional layer. Our weights were also pre-initialized using weights from the neural network trained to predict objects in the ImageNet data set \cite{Russakovsky2015}.

The neural network in the original study was trained to output multiple binary predictions: 1) whether the image was graded moderate or worse diabetic retinopathy (i.e. moderate, severe, or proliferative grades); 2) severe or worse diabetic retinopathy; 3) referable diabetic macular edema; or 4) fully gradable. The term referable diabetic retinopathy was defined in the original study as an image associated with either or both category 1) and 3). For the training data obtained in this replication study, only grades for diabetic retinopathy were present. That means that our neural network outputs only one binary prediction: moderate or worse diabetic retinopathy (referable diabetic retinopathy).

In this study, the training and validation sets were split like in the original study: 80\% was used for training and 20\% was used for validating the neural network. It is estimated that 25\% of the Kaggle EyePACS set consists of ungradable images \cite{Rakhlin2017}. Therefore, we also assessed image gradability for all Kaggle EyePACS images, and we trained an algorithm with only gradable images. In the original study, the performance of an algorithm trained with only gradable images was also summarized. We do not use the image quality grades as an input for algorithm training.

Specific details on the image normalization method or hyper-parameters of the validation procedure were not specified, so we conducted experiments to find the normalization method and hyper-parameter settings that worked well for training and validating the algorithms. We trained with three normalization methods: 1) image standardization, which involves subtracting the mean from each image and dividing each image by the standard deviation; 2) normalizing images to a [0, 1] range; and 4) normalizing images to a [--1, 1] range.

\subsection{Algorithm validation}

We validate the algorithm by measuring the performance of the resulting neural network by the area under the receiver operating characteristic curve (AUC) on a validation set, as in the original study. We find the area by thresholding the network’s output predictions, which are continuous numbers ranging from 0 to 1. By moving the operating threshold on the predictions, we obtain different results for sensitivity and specificity. We then plot sensitivity against 1–specificity for 200 thresholds. Finally, the AUC of the validation set is calculated, and becomes an indicator for how well the neural network detects referable diabetic retinopathy. The original study did not describe how many thresholds were used for plotting AUC, so we used the de facto standard of 200 thresholds.

The original paper describes that the AUC value of the validation set was used for the early-stopping criterion \cite{Caruana2000}; training is terminated when a peak AUC on the validation set is reached. This prevents overfitting the neural network on the training set. In our validation procedure, we also use the AUC calculated from the validation set as an early stopping criterion. To determine if a peak AUC is reached, we compared the AUC values between different validation checkpoints. To avoid stopping at a local maximum of the validation AUC function, our network may continue to perform training up to n epochs (i.e. patience of n epochs). Since the original paper did not describe details regarding the validation procedure, we had to experiment with several settings for patience. One epoch of training is equal to running all images through the network once.

We used ensemble learning \cite{Krizhevsky2012} by training 10 networks on the same data set, and using the final prediction computed by taking the mean of the predictions of the ensemble. This was also done in the original study.

In the original study, additional experiments were conducted to evaluate the performance of the resulting algorithm based on the training set, compared with performance based on subsets of images and grades from the training set. We did not replicate these experiments for two reasons. First, we chose to focus on replicating the main results of the original paper. That is, the results of an algorithm detecting referable diabetic retinopathy. Second, we cannot perform subsampling of grades, as we only have one grade per image.

\section{Results}

As for the early-stopping criterion at a peak AUC, we found that a patience of 10 epochs worked well. Our chosen requirement for a new peak AUC was a value of AUC that is larger than the previous peak value, with a minimum difference of 0.01. The normalization method of normalizing the images to a [--1, 1] range outperformed the other normalization methods.

\begin{figure}
  \centering
    \includegraphics[width=\textwidth]{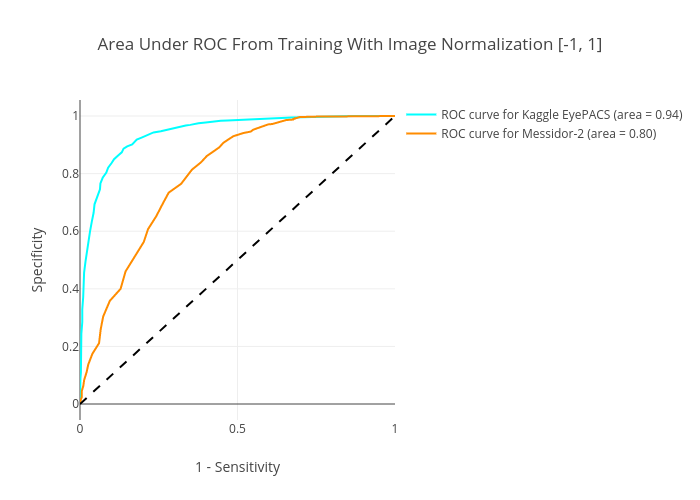}
  \caption{Area under receiver operating characteristic curve for training with normalizing images to a [--1, 1] range.\label{fig:auc-1-to-1}}
\end{figure}

\begin{figure}
  \centering
    \includegraphics[width=\textwidth]{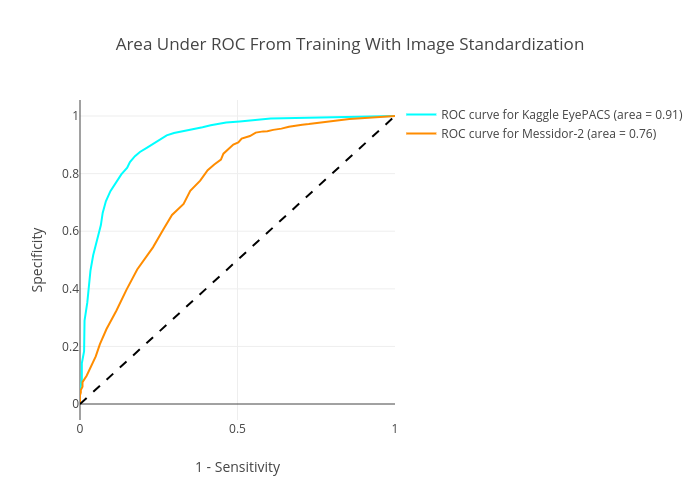}
  \caption{Area under receiver operating characteristic curve for training with standardizing images.\label{fig:auc-standardization}}
\end{figure}

\begin{figure}
  \centering
    \includegraphics[width=\textwidth]{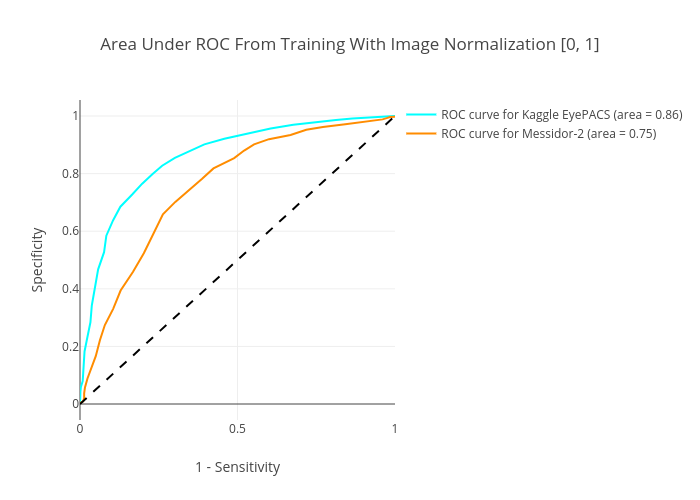}
  \caption{Area under receiver operating characteristic curve for training with normalizing images to a [0, 1] range.\label{fig:auc-0-to-1}}
\end{figure}

The replica algorithm's performance was evaluated on two independent test sets. We provide an overview of the differences in image distribution used in our replication study compared with the original study in Figure~\ref{fig:data-distribution} in Section~\ref{sec:data-grading}. Our replica algorithm trained with normalizing images to a [--1, 1] range yielded an AUC of 0.94 and 0.80 on our Kaggle EyePACS test data set and Messidor-2, respectively (Figure~\ref{fig:auc-1-to-1} and Table~\ref{table:results}). We observe that there is a large discrepancy between the AUC of our replication study and the original study. Our results for training with the other conducted normalization methods are also shown in Table~\ref{table:results}. Figure~\ref{fig:auc-standardization} and~\ref{fig:auc-0-to-1} show their corresponding receiver operating characteristic curves. Lastly, we attempted training by excluding non-gradable images, but this has shown to not increase algorithm performance.

\newgeometry{bottom=1cm,top=1cm,footskip=1em,includefoot}

\clearpage

\restoregeometry

\begin{table}[ht]
\centering
\resizebox{\textwidth}{!}{%
\begin{tabular}{@{}llll@{}}

\multicolumn{4}{c}{\textbf{Replication results}} 
\\ \multicolumn{4}{l}{}      \\
\multicolumn{4}{c}{\textbf{Normalizing images to [--1, 1] range}}                                                                                                                                                                                                                                                          \\ \midrule \\
\textit{\textbf{Test set}}                                                       & \textit{\textbf{High sensitivity}}                                                      & \textit{\textbf{High specificity}}                                                      & \textit{\textbf{AUC score}}         
\\ \\
\begin{tabular}[c]{@{}l@{}}Kaggle EyePACS test \\ (orig. EyePACS-1)\end{tabular} & \begin{tabular}[c]{@{}l@{}}89.9 (97.5)\% sens.\\ 83.8 (93.4)\% spec.\end{tabular} & \begin{tabular}[c]{@{}l@{}}83.4 (90.3)\% sens.\\ 90.1 (98.1)\% spec.\end{tabular} & 0.94 (0.99)                        \\ \\
\begin{tabular}[c]{@{}l@{}}Messidor-2\end{tabular}  & \begin{tabular}[c]{@{}l@{}}73.7 (96.1)\% sens.\\ 69.7 (93.9)\% spec.\end{tabular} & \begin{tabular}[c]{@{}l@{}}67.9 (87.0)\% sens.\\ 76.4 (98.5)\% spec.\end{tabular} & 0.80 (0.99)                        \\
\multicolumn{4}{l}{}                                                                                                                                                                                                                                                                                      \\
\multicolumn{4}{c}{\textbf{Image standardization}}                                                                                                                                                                                                                                                          \\ \midrule \\
\textit{\textbf{Test set}}                                                       & \textit{\textbf{High sensitivity}}                                                      & \textit{\textbf{High specificity}}                                                      & \textit{\textbf{AUC score}}         
\\ \\
\begin{tabular}[c]{@{}l@{}}Kaggle EyePACS test \\ (orig. EyePACS-1)\end{tabular} & \begin{tabular}[c]{@{}l@{}}88.3 (97.5)\% sens.\\ 77.1 (93.4)\% spec.\end{tabular} & \begin{tabular}[c]{@{}l@{}}78.8 (90.3)\% sens.\\ 88.9 (98.1)\% spec.\end{tabular} & 0.91 (0.99)                        \\ \\
\begin{tabular}[c]{@{}l@{}}Messidor-2\end{tabular}  & \begin{tabular}[c]{@{}l@{}}73.4 (96.1)\% sens.\\ 60.9 (93.9)\% spec.\end{tabular} & \begin{tabular}[c]{@{}l@{}}65.0 (87.0)\% sens.\\ 74.1 (98.5)\% spec.\end{tabular} & 0.76 (0.99)                        \\
\multicolumn{4}{l}{}      \\
\multicolumn{4}{c}{\textbf{Normalizing images to [0, 1] range}}                                                                                                                                                                                                                                                          \\ \midrule \\
\textit{\textbf{Test set}}                                                       & \textit{\textbf{High sensitivity}}                                                      & \textit{\textbf{High specificity}}                                                      & \textit{\textbf{AUC score}}         
\\ \\
\begin{tabular}[c]{@{}l@{}}Kaggle EyePACS test \\ (orig. EyePACS-1)\end{tabular} & \begin{tabular}[c]{@{}l@{}}83.4 (97.5)\% sens.\\ 72.7 (93.4)\% spec.\end{tabular} & \begin{tabular}[c]{@{}l@{}}73.9 (90.3)\% sens.\\ 82.7 (98.1)\% spec.\end{tabular} & 0.86 (0.99)                        \\ \\
\begin{tabular}[c]{@{}l@{}}Messidor-2\end{tabular}  & \begin{tabular}[c]{@{}l@{}}73.7 (96.1)\% sens.\\ 65.9 (93.9)\% spec.\end{tabular} & \begin{tabular}[c]{@{}l@{}}64.5 (87.0)\% sens.\\ 75.1 (98.5)\% spec.\end{tabular} & 0.75 (0.99)                        \\
\multicolumn{4}{l}{} \\ \bottomrule 
\end{tabular}%
}
\caption{Performance on test sets of replication with various normalization methods, compared to results from the original study. The results of the original study are depicted in parenthesizes.}
\label{table:results}
\end{table}

\clearpage

\section{Discussion}

The results show substantial performance differences between the original study’s algorithm and our replica algorithm. Even though we followed the methodology of the original study as closely as possible, our algorithm did not come close to the results in the original study. This is probably because our algorithms were trained with different public data, under different hyper-parameters, and because in the original study ophthalmologic experts re-graded all their images. According to the original study, the validation and test sets should have multiple grades per image, because it will provide a more reliable measure of a model’s final predictive ability. Their results on experimenting with only one grade per image show that their algorithm’s performance declines with 36\%.

The hyper-parameters were not published when we started this replication study. Later, hyper-parameters for training and data augmentation were published in \cite{Krause2018}, and then we retrained all algorithms with these hyper-parameters and data augmentation settings. However, some of the details for the methods in the original study remain unspecified. First, the hyper-parameter settings for the validation procedure and the used normalization method are missing. Second, it is unclear how the algorithm’s predictions for diabetic retinopathy or macular edema are interpreted in case of ungradable images. The image quality grades might have been used as an input for the network, or the network might be concatenated with another network that takes the image quality as an input. Third, apart from the main algorithm that detects referable diabetic retinopathy and outputs 4 binary classifications, other algorithms seem to have been trained as well. An example is the described algorithm that only detects referable diabetic retinopathy for gradable images, and an algorithm that detects all-cause referable diabetic retinopathy, which presents moderate or worse diabetic retinopathy, referable macular edema, and ungradable images. Details on how these other algorithms are built are however not reported. It is unclear whether the main network has been used or if the original study trained new networks. Lastly, the original paper did not state how many iterations it took for their proposed model to converge during training, or describe how to find a converging model.

Our results show that picking the appropriate normalization method is essential. From the three different methods we trained with, image normalization to a [--1, 1] range turned out to be the best performing method and is therefore assumed to be used in the original study. This is also likely due to the fact that the pre-trained InceptionV3 network was trained with [--1, 1] normalized ImageNet images.

\subsection{Hyper-parameters}

The main challenge in this replication study was to find hyper-parameters, which were not specified in the original paper, such that the algorithm does not converge on a local maximum of the validation AUC function. To understand how we should adjust the hyper-parameters, we measured the Brier score on the training set and the AUC value on the validation set after each epoch of training. One possible reason for the algorithm having problems to converge may be the dimensions of the fundus images. As the original study suggests, the original fundus images were preprocessed and scaled down to a width and height of 299 pixels to be able to initialize the InceptionV3 network with ImageNet pre-trained weights, which have been trained with images of 299 by 299 pixels. We believe it is difficult for ophthalmologists to find lesions in fundus images of this size, so we assume the algorithm has difficulties with detecting lesions as well. \cite{Rakhlin2017} also points out this fact, and suggests re-training an entire network with larger fundus images and randomly initialized weights instead. And as mentioned before, it seems like the original study extended the InceptionV3 model architecture for their algorithm to use image gradability as an input parameter.

\subsection{Kaggle images}

A potential drawback with the images from Kaggle is that it contains grades for diabetic retinopathy for all images. We found that 19.9\% of these images is ungradable, and it is thus possible that the algorithm will “learn” features for ungradable images, and make predictions based on anomalies. This is likely to negatively contribute to the algorithm’s predictive performance, but we were not able to show a significant difference of performance between an algorithm trained on all images and an algorithm trained on only gradable images. 

\section{Conclusion}

We  re-implemented  the  main  method  from  JAMA  2016;  316(22),  but  we  were  not  able  to  get the same performance as reported in that study using publicly available data. The main identified sources for deviation between the original and the replicated results are hyper-parameters, and quality of data and grading.
From trying several normalization methods, we found that the original study most likely normalized images to a [--1, 1] range, because it yielded the best performing algorithm, but its results still deviate from the original algorithm's results.
We assume the impact of the hyper-parameters to be minor, but there is no standard setting for hyper-parameters, and we therefore regard this as missing level of detail. 
The original study had access to data of higher quality than those that are publicly available, and this is likely to account for part of the deviation in results. Gulshan et al showed that the performance levels off around 40 000 (Figure 4A), and we therefore assume that the reduced size of replication data set is not a large source for deviation in the results. 
The number of grades per image is a possible explanation of Gulshan et al's superior results, but the impact is uncertain. Figure 4B depicts performance as a function of grades, but there is an overfitting component: 100\% vs. 65\% specificity for the training and test set, respectively, and it is not possible to distinguish the contribution from the overfitting from that of the low number of grades. 

The source code of this replication study and instructions for running the replication are available at \url{https://github.com/mikevoets/jama16-retina-replication}.

\newpage

\bibliographystyle{myIEEEtran}
\bibliography{IEEEabrv,mendeley}

\begin{thebibliography}{10}
\providecommand{\url}[1]{#1}
\csname url@samestyle\endcsname
\providecommand{\newblock}{\relax}
\providecommand{\bibinfo}[2]{#2}
\providecommand{\BIBentrySTDinterwordspacing}{\spaceskip=0pt\relax}
\providecommand{\BIBentryALTinterwordstretchfactor}{4}
\providecommand{\BIBentryALTinterwordspacing}{\spaceskip=\fontdimen2\font plus
\BIBentryALTinterwordstretchfactor\fontdimen3\font minus
  \fontdimen4\font\relax}
\providecommand{\BIBforeignlanguage}[2]{{%
\expandafter\ifx\csname l@#1\endcsname\relax
\typeout{** WARNING: IEEEtran.bst: No hyphenation pattern has been}%
\typeout{** loaded for the language `#1'. Using the pattern for}%
\typeout{** the default language instead.}%
\else
\language=\csname l@#1\endcsname
\fi
#2}}
\providecommand{\BIBdecl}{\relax}
\BIBdecl

\bibitem{Gulshan2016}
V.~Gulshan, L.~Peng, M.~Coram, M.~C. Stumpe, D.~Wu, A.~Narayanaswamy,
  S.~Venugopalan, K.~Widner, T.~Madams, J.~Cuadros, R.~Kim, R.~Raman, P.~C.
  Nelson, J.~L. Mega, and D.~R. Webster, ``{Development and validation of a
  deep learning algorithm for detection of diabetic retinopathy in retinal
  fundus photographs},'' \emph{JAMA - Journal of the American Medical
  Association}, vol. 316, no.~22, pp. 2402--2410, 2016. doi:
  10.1001/jama.2016.17216

\bibitem{eLife:20176}
``{The challenges of replication},'' \emph{eLife}, vol.~6, no. e23693, 2017.
  doi: 10.7554/eLife.23693

\bibitem{Nature:496}
``{Announcement: Reducing our irreproducibility},'' \emph{Nature}, vol. 496,
  no. 7446, pp. 398--398, 2013. doi: 10.1038/496398a

\bibitem{Nature:533}
``{Reality check on reproducibility},'' \emph{Nature}, vol. 533, no. 7604, p.
  437, 2016. doi: 10.1038/533437a

\bibitem{Hutson2018}
\BIBentryALTinterwordspacing
M.~Hutson, ``{Missing data hinder replication of artificial intelligence
  studies},'' 2018. [Online]. Available:
  \url{http://www.sciencemag.org/news/2018/02/missing-data-hinder-replication-artificial-intelligence-studies}
\BIBentrySTDinterwordspacing

\bibitem{Litjens2017}
G.~Litjens, T.~Kooi, B.~E. Bejnordi, A.~A.~A. Setio, F.~Ciompi, M.~Ghafoorian,
  J.~A. van~der Laak, B.~van Ginneken, and C.~I. S{\'{a}}nchez, ``{A survey on
  deep learning in medical image analysis},'' pp. 60--88, 2017.

\bibitem{Ching2018}
T.~Ching, D.~S. Himmelstein, B.~K. Beaulieu-Jones, A.~A. Kalinin, B.~T. Do,
  G.~P. Way, E.~Ferrero, P.-M. Agapow, M.~Zietz, M.~M. Hoffman, W.~Xie, G.~L.
  Rosen, B.~J. Lengerich, J.~Israeli, J.~Lanchantin, S.~Woloszynek, A.~E.
  Carpenter, A.~Shrikumar, J.~Xu, E.~M. Cofer, C.~A. Lavender, S.~C. Turaga,
  A.~M. Alexandari, Z.~Lu, D.~J. Harris, D.~DeCaprio, Y.~Qi, A.~Kundaje,
  Y.~Peng, L.~K. Wiley, M.~H.~S. Segler, S.~M. Boca, S.~J. Swamidass, A.~Huang,
  A.~Gitter, and C.~S. Greene, ``{Opportunities and obstacles for deep learning
  in biology and medicine},'' \emph{Journal of The Royal Society Interface},
  vol.~15, no. 141, 2018. doi: 10.1098/rsif.2017.0387

\bibitem{GoogleScholar:Citations}
\BIBentryALTinterwordspacing
{Google}, ``{Citations for Development and Validation of a Deep Learning
  Algorithm for Detection of Diabetic Retinopathy in Retinal Fundus
  Photographs},'' 2018. [Online]. Available:
  \url{https://scholar.google.no/scholar?cites=16083985573643781536}
\BIBentrySTDinterwordspacing

\bibitem{Ting2017DevelopmentDiabetes}
D.~S.~W. Ting, C.~Y.-L. Cheung, G.~Lim, G.~S.~W. Tan, N.~D. Quang, A.~Gan,
  H.~Hamzah, R.~Garcia-Franco, I.~Y. San~Yeo, S.~Y. Lee, E.~Y.~M. Wong,
  C.~Sabanayagam, M.~Baskaran, F.~Ibrahim, N.~C. Tan, E.~A. Finkelstein, E.~L.
  Lamoureux, I.~Y. Wong, N.~M. Bressler, S.~Sivaprasad, R.~Varma, J.~B. Jonas,
  M.~G. He, C.-Y. Cheng, G.~C.~M. Cheung, T.~Aung, W.~Hsu, M.~L. Lee, and T.~Y.
  Wong, ``{Development and Validation of a Deep Learning System for Diabetic
  Retinopathy and Related Eye Diseases Using Retinal Images From Multiethnic
  Populations With Diabetes},'' \emph{JAMA}, vol. 318, no.~22, p. 2211, 2017.
  doi: 10.1001/jama.2017.18152

\bibitem{Abramoff2016ImprovedLearning}
M.~D. Abr{\`{a}}moff, Y.~Lou, A.~Erginay, W.~Clarida, R.~Amelon, J.~C. Folk,
  and M.~Niemeijer, ``{Improved automated detection of diabetic retinopathy on
  a publicly available dataset through integration of deep learning},''
  \emph{Investigative Ophthalmology and Visual Science}, vol.~57, no.~13, pp.
  5200--5206, 2016. doi: 10.1167/iovs.16-19964

\bibitem{Gargeya2017AutomatedLearning}
R.~Gargeya and T.~Leng, ``{Automated Identification of Diabetic Retinopathy
  Using Deep Learning},'' \emph{Ophthalmology}, vol. 124, no.~7, pp. 962--969,
  2017. doi: 10.1016/j.ophtha.2017.02.008

\bibitem{Krause2018}
J.~Krause, V.~Gulshan, E.~Rahimy, P.~Karth, K.~Widner, G.~S. Corrado, L.~Peng,
  and D.~R. Webster, ``Grader variability and the importance of reference
  standards for evaluating machine learning models for diabetic retinopathy,''
  \emph{arXiv:1710.01711 [cs.CV]}, 2017.

\bibitem{Decenciere2014}
E.~Decenci{\`{e}}re, X.~Zhang, G.~Cazuguel, B.~Lay, B.~Cochener, C.~Trone,
  P.~Gain, R.~Ordonez, P.~Massin, A.~Erginay, B.~Charton, and J.-C. Klein,
  ``{Feedback on a publicly distributed image database: the Messidor
  database},'' \emph{Image Analysis {\&} Stereology}, vol.~33, no.~3, p. 231,
  2014. doi: 10.5566/ias.1155

\bibitem{Quellec2008}
G.~Quellec, M.~Lamard, P.~M. Josselin, G.~Cazuguel, B.~Cochener, and C.~Roux,
  ``{Optimal Wavelet Transform for the Detection of Microaneurysms in Retina
  Photographs},'' \emph{IEEE Transactions on Medical Imaging}, vol.~27, no.~9,
  pp. 1230--1241, 2008. doi: 10.1109/TMI.2008.920619

\bibitem{Kaggle:Main}
\BIBentryALTinterwordspacing
{Kaggle}, ``{Diabetic Retinopathy Detection (Data)},'' 2015. [Online].
  Available: \url{https://www.kaggle.com/c/diabetic-retinopathy-detection/data}
\BIBentrySTDinterwordspacing

\bibitem{Kaggle:Data}
\BIBentryALTinterwordspacing
------, ``{Diabetic Retinopathy Detection},'' 2015. [Online]. Available:
  \url{https://www.kaggle.com/c/diabetic-retinopathy-detection}
\BIBentrySTDinterwordspacing

\bibitem{Rakhlin2017}
A.~Rakhlin, ``{Diabetic Retinopathy detection through integration of Deep
  Learning classification framework},'' \emph{bioRxiv}, 2017. doi:
  10.1101/225508

\bibitem{Icoph:2010}
C.~Wilkinson, F.~L. Ferris, R.~E. Klein, P.~P. Lee, C.~D. Agardh, M.~Davis,
  D.~Dills, A.~Kampik, R.~Pararajasegaram, J.~T. Verdaguer, and {Global
  Diabetic Retinopathy Project Group}, ``{Proposed international clinical
  diabetic retinopathy and diabetic macular edema disease severity scales},''
  \emph{Ophthalmology}, vol. 110, no.~9, pp. 1677--1682, 2003. doi:
  10.1016/S0161-6420(03)00475-5

\bibitem{AbramoffDRGrades}
\BIBentryALTinterwordspacing
M.~D. Abramoff, ``{Abramoff Messidor-2 reference standard rDR 20 July 2016}.''
  [Online]. Available:
  \url{http://webeye.ophth.uiowa.edu/abramoff/abramoff-messidor-2-reference%20standard%20rDR%20july%202016.pdf}
\BIBentrySTDinterwordspacing

\bibitem{Szegedy2015}
C.~Szegedy, V.~Vanhoucke, S.~Ioffe, J.~Shlens, and Z.~Wojna, ``{Rethinking the
  Inception Architecture for Computer Vision},'' \emph{arXiv:1512.00567
  [cs.CV]}, 2015. doi: 10.1109/CVPR.2016.308

\bibitem{Dean2012}
J.~Dean, G.~Corrado, R.~Monga, K.~Chen, M.~Devin, M.~Mao, M.~Ranzato,
  A.~Senior, P.~Tucker, K.~Yang, Q.~V. Le, and A.~Y. Ng, ``{Large Scale
  Distributed Deep Networks},'' in \emph{Advances in Neural Information
  Processing Systems 25 (NIPS 25)}.\hskip 1em plus 0.5em minus 0.4em\relax
  Curran Associates Inc., 2012, pp. 1223--1231.

\bibitem{Ioffe2015}
S.~Ioffe and C.~Szegedy, ``{Batch Normalization: Accelerating Deep Network
  Training by Reducing Internal Covariate Shift},'' \emph{arXiv:1502.03167
  [cs.LG]}, 2015. doi: 10.1007/s13398-014-0173-7.2

\bibitem{Russakovsky2015}
O.~Russakovsky, J.~Deng, H.~Su, J.~Krause, S.~Satheesh, S.~Ma, Z.~Huang,
  A.~Karpathy, A.~Khosla, M.~Bernstein, A.~C. Berg, and L.~Fei-Fei, ``{ImageNet
  Large Scale Visual Recognition Challenge},'' \emph{International Journal of
  Computer Vision}, vol. 115, no.~3, pp. 211--252, 2015. doi:
  10.1007/s11263-015-0816-y

\bibitem{Caruana2000}
R.~Caruana, S.~Lawrence, and L.~Giles, ``{Overfitting in neural nets:
  Backpropagation, conjugate gradient, and early stopping},'' in \emph{the 13th
  International Conference on Neural Information Processing Systems}.\hskip 1em
  plus 0.5em minus 0.4em\relax Advances in Neural Information Processing
  Systems 13 (NIPS 2000), 2000. doi: 10.1109/IJCNN.2000.857823. ISBN 1049-5258.
  ISSN 10495258 pp. 402--408.

\bibitem{Krizhevsky2012}
A.~Krizhevsky, I.~Sutskever, and G.~E. Hinton, ``{ImageNet Classification with
  Deep Convolutional Neural Networks},'' in \emph{Advances In Neural
  Information Processing Systems}.\hskip 1em plus 0.5em minus 0.4em\relax
  Advances in Neural Information Processing Systems 25 (NIPS 2012), 2012. doi:
  http://dx.doi.org/10.1016/j.protcy.2014.09.007. ISBN 9781627480031. ISSN
  10495258 pp. 1--9.

\end{thebibliography}

\end{document}